\documentclass[letterpaper, 10 pt, conference]{ieeeconf}  % Comment this line out if you need a4paper

\IEEEoverridecommandlockouts                              % This command is only needed if 
\newcommand{\algabbr}{Lifelong LERF}
\newcommand{\algname}{Lifelong LERF}

\usepackage{graphics}
\usepackage[pdftex]{graphicx}
\usepackage{wrapfig}
\usepackage{enumerate}   
\DeclareGraphicsExtensions{.pdf,.png,.jpg}
\usepackage{epsfig}
\usepackage[font={small}]{caption}
\usepackage{subcaption}
\usepackage[rightcaption]{sidecap}
\usepackage{pbox}
\usepackage{bigstrut}
\usepackage{makecell}
\usepackage{ctable} % for \specialrule command

\usepackage{amssymb,amsmath}
\usepackage{gensymb} 
\usepackage{nicefrac}       % compact symbols for 1/2, etc.
\numberwithin{equation}{section} 
\usepackage{algorithm}
\usepackage{algpseudocode}

\usepackage{textcomp} %for getting text tilde
\usepackage{array} %for table entries to be in center of cell
\usepackage{tabularx}
\usepackage{multirow}
\usepackage{multicol}
\usepackage{booktabs}
\usepackage{tabulary}

\usepackage[utf8]{inputenc}
\usepackage{units}
\usepackage{bm}
\usepackage{xspace}
\usepackage{flushend}%balance columns on last page
\usepackage{balance} 
\usepackage{csquotes}
\usepackage{makeidx}
\usepackage{blindtext}
\usepackage{xcolor}
\usepackage{caption}
\usepackage{subcaption}

\usepackage{url}
\usepackage{hyperref}
\hypersetup{
    colorlinks=true,
    linkcolor=black,
    citecolor=black,
    filecolor=cyan,
    urlcolor=black
}

\usepackage[backend=biber,
            url=false,
            isbn=false,
            doi=false,
            backref=false,
            style=ieee,
            natbib=true,%compatibility aliases
            mincitenames=1,
            maxcitenames=1,
            citestyle=numeric,
            sorting=none,%none
            block=none]{biblatex}
\renewcommand{\bibfont}{\small}
\renewcommand{\baselinestretch}{1.0}



\title{\LARGE \bf
Lifelong LERF: Local 3D Semantic Inventory Monitoring \\Using FogROS2 
}

\author{Adam Rashid$^{1}$*, Chung Min Kim$^{1}$*, Justin Kerr$^{1}$*, Letian Fu$^{1}$, Kush Hari$^{1}$, Ayah Ahmad$^{1}$, Kaiyuan Chen$^{2}$, \\Huang Huang$^{1}$, Marcus Gualtieri$^{3}$, Michael Wang$^{3}$, Christian Juette$^{3}$, Nan Tian$^{3}$, Liu Ren$^{3}$, Ken Goldberg$^{1}$
\thanks{*Equal contribution, $^{1}$The AUTOLab at $^{2}$UC Berkeley, $^{3}$Bosch}%
}

\begin{document}

\maketitle
\thispagestyle{empty}
\pagestyle{empty}

%%%%%%%%%%%%%%%%%%%%%%%%%%%%%%%%%%%
\begin{abstract}
% Many environments require a navigating robot to monitor for extensive periods, during which objects in the scene may shift significantly. I
Inventory monitoring in homes, factories, and retail stores relies on maintaining data despite objects being swapped, added, removed, or moved. We introduce \algname{}, a method that allows a mobile robot with minimal compute to jointly optimize a dense language and geometric representation of its surroundings. \algname{} maintains this representation over time by detecting semantic changes and selectively updating these regions of the environment, avoiding the need to exhaustively remap. Human users can query inventory by providing natural language queries and receiving a 3D heatmap of potential object locations. To manage the computational load, we use Fog-ROS2, a cloud robotics platform, to offload resource-intensive tasks. \algname{} obtains poses from a monocular RGBD SLAM backend, and uses these poses to progressively optimize a Language Embedded Radiance Field (LERF) for semantic monitoring. Experiments with 3-5 objects arranged on a tabletop and a Turtlebot with a RealSense camera suggest that \algabbr{} can persistently adapt to changes in objects with up to 91\% accuracy.

\end{abstract}

\section{Introduction}
% Outline: 
% 1. Introduce semantic scene reconstruction from the perspective of scene reconstruction (SLAM, NeRF), why NeRF/LeRF is preferable compared to traditional methods (cite LeRF for conclusion) 
% 2. Explain why lifelong learning is important for scene reconstruction (scene updates over time, Plenoptic Function), and most of the current frameworks do not actively detect differences between previously recorded states and the current state, leading to computation overhead. 
% 3. Introduce LifeLong LeRF, an active SLAM + LeRF framework that builds and updates a LeRF based on changes detected in the scene, with a list of contributions: 
% a. Algorithmic contribution: unsupervised scene-change detection via Dino attention map or CLIP features, which can be either class-agnostic or class-aware (query dependent). 
% b. Framework: Fog-Rosify both LeRF and DROID-SLAM, thereby leveraging cloud computing for real-time mapping and updating of LeRF.
% c. Resulting algorithm successfully detects changes in the environment correctly x percent of the time.

The ability to monitor inventory over time and respond to natural language queries such as ``where did I leave my keys?" or ``where is the yellow electric screwdriver?" can be useful in homes, retail shops, repair shops, hospitals, and factories. Indoor scenes evolve over time, with objects moving, changing, appearing, or disappearing. To remain useful, systems must keep track of the current environment state. We consider how a mobile robot (wheeled or legged) can navigate using Simultaneous Localization and Mapping (SLAM) ~\cite{schoenberger2016sfm, mur2015orb, teed2021droid} to collect images, build, query, and update semantic 3D models over time. 

\algname{} enables a mobile robot with limited on-board compute to \textit{create} a dense semantic representation that supports natural language queries, and \textit{maintain} this representation by automatically detecting semantic differences in 3D as it navigates, updating the reconstruction to reflect these changes. \algabbr{} uses DROID-SLAM as its monocular pose estimation backend, and uses these poses to progressively optimize a Language Embedded Radiance Field~\cite{lerf2023} (LERF) which densely embeds CLIP~\cite{radford2021learning} embeddings within a Neural Radiance Field (NeRF)~\cite{mildenhall2021nerf}. 

% SLAM algorithms allow robots to localize themselves while building a map of their surroundings. SLAM primarily focuses on geometric reconstruction; building 3D maps such as pointclouds or TSDFs. Recent variants provide impressive tracking and reconstruction abilities. In addition, several recent works augment these 3D maps with semantics, either through pre-trained object detectors~\cite{conceptfusion}, or vision-language encoders~\cite{lerf2023, kobayashi2022distilledfeaturefields, tschernezki22neural} to support higher-level planning and decision-making.

% \begin{figure}[t]
% \centering
% \includegraphics[width=\linewidth]{figures/splash_proglerf.pdf}
% \caption{\textit{\algname{} Update Example}. \textbf{Top}: A mobile robot takes a scan of the scene and builds a Language Embedded Radiance Field (LERF) that reconstructs the scene. As an example, the ``Clorox" cleaning wipes are replaced by a cookie can. The robot rescans the scene and provides new captures of the scene. \textbf{Middle}: To identify what has changed, we render the semantic features stored in LERF and compare them against those extracted from the newly captured images. We then de-project the changes into 3D bounding boxes. \textbf{Bottom}: LERF updates the reconstruction inside the 3D bounding boxes using new images of the scene, progressively changing the geometry and semantics.}
% \end{figure}
\begin{figure}
\centering
\includegraphics[width=\linewidth]{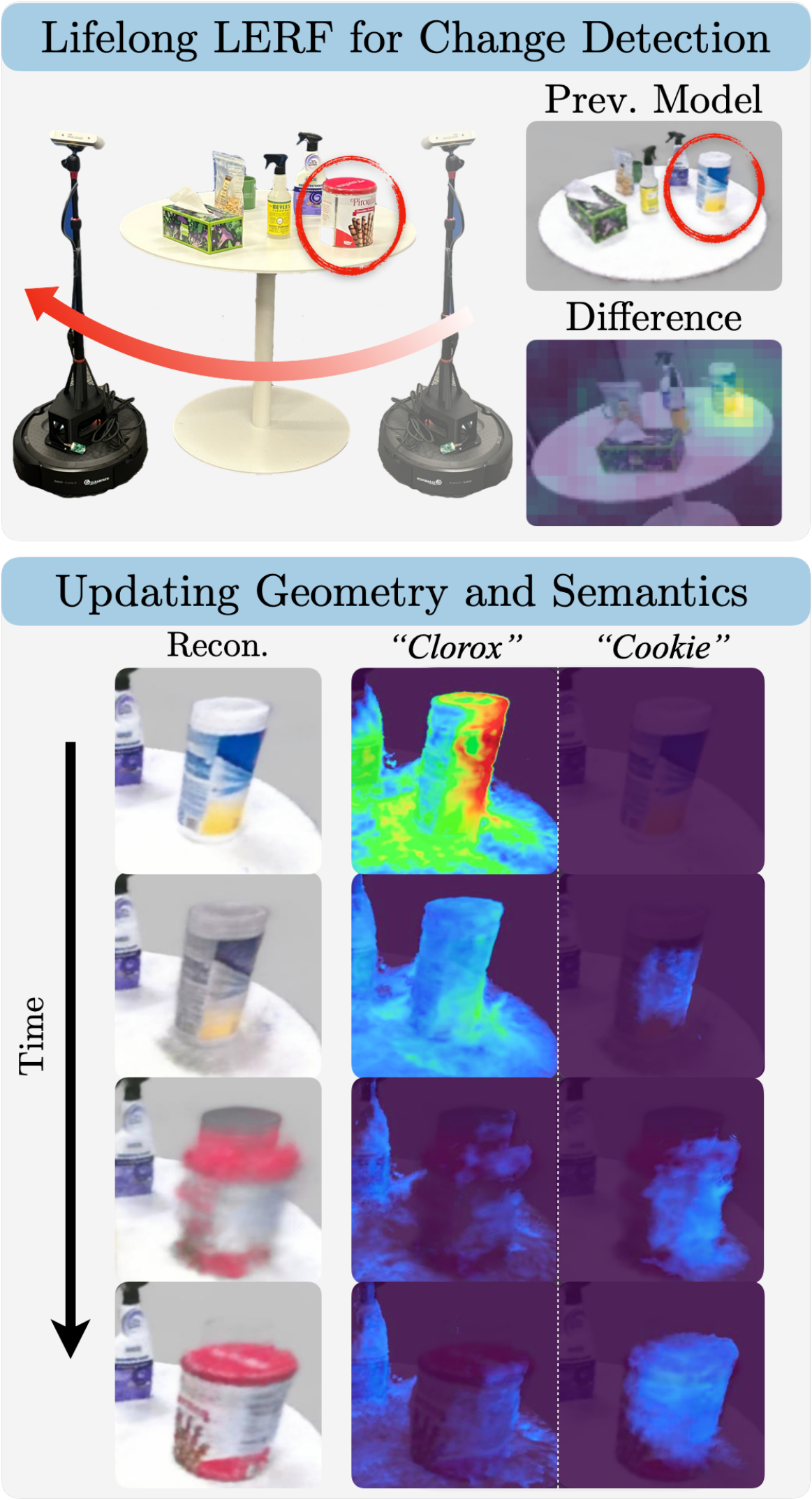}
\caption{\textit{\algname{} Example}. \textbf{Top}: A mobile robot takes a scan of the scene and builds a Language Embedded Radiance Field (LERF). Then, the scene is altered, for example the ``Clorox" wipes are replaced by a cookie can. The robot periodically rescans the scene and identifies what has changed by rendering semantic features stored in LERF and comparing them against those extracted from the newly captured images. \textbf{Bottom}: Lifelong LERF efficiently updates the LERF using new images of the scene, progressively changing local geometry and semantics.}
\end{figure}

Prior work Evo-NeRF~\cite{kerr2022evo} demonstrates NeRFs can update a scene by discarding stale images and replacing them with fresh images, similar to how many SLAM models can update an occupancy distribution. However, for inventory tracking it is impractical to throw away old images and remap the entire environment. In this work, we present a new method for localizing changed regions in an environment to selectively update. 

\algabbr{} detects 3D differences based on the deviation between natural language embeddings computed over a freshly captured image and a view previously rendered from the LERF at the same pose, leveraging the novel view synthesis capabilities of LERF to generate pixel-aligned difference heatmaps. Given these scene changes, the robot can update the LERF by masking out stale regions of previous input images and ignoring these pixels during future ray sampling. As fresh images are collected to the updated scene, the underlying LERF evolves to match new observations.

The compute required for running SLAM, optimizing a LERF, and computing semantic differences is too demanding for on-board robot hardware. We address this issue with cloud computing by using FogROS2~\cite{chen2023fogros}. We conduct experiments on a Turtlebot 4 and evaluate \algabbr{}'s ability to maintain an updated language representation as changes are made to local tabletop scenes. We find that experiments in a tabletop environment suggest \algabbr{} can persistently localize objects through scene changes up to 91\% of the time, with the proposed semantic difference method outperforming a depth baseline in robustness to false positives and object swapping. This method could be complementary to methods using semantics for downstream applications such as language-conditioned grasping~\cite{lerftogo2023}.

This paper contributes:
\begin{enumerate}
    \item A novel method for detecting changes in a scene by comparing rendered language embeddings against those calculated from captured images.
    \item An approach for incorporating detected scene changes progressively into an evolving LERF.
    \item A system that allows \algabbr{} to function on a robot with lightweight compute (Turtlebot 4 with Raspberry Pi 4) by leveraging FogROS2.
    % \item Experiments suggesting \algabbr{} successfully detects changes in tabletop environments correctly \todo{} percent of the time.
\end{enumerate}

%This paper contributes: 1) A novel method for detecting inventory changes in a scene by comparing rendered language embeddings against those calculated from captured images. 2) An approach for incorporating detected scene changes progressively into a LERF. 3) A cloud architecture that allows \algabbr{} to function on a robot with lightweight compute (Turtlebot 4 with Raspberry Pi 4) in real time. 4) Experiments suggesting \algabbr{} successfully detects changes in tabletop environments correctly \todo{} percent of the time.%

    % \item We introduce an unsupervised scene-change detection mechanism using DINO attention maps or CLIP features. This can be configured to be either class-agnostic or class-aware depending on the specific queries.
    % \item We extend the capabilities of both LeRF and DROID-SLAM by integrating them with FogROS. This allows us to leverage cloud computing resources for real-time mapping and updating of LeRF, making our framework well-suited for resource-constrained robotic systems. %P1
\section{Related Work}\label{sec:related_work}
% \subsection{Semantic Scene Models} \todo{Break up into 2 sections, SLAM and Dynamic/Semantic SLAM}

% TODO merge the following to visual salm 

\subsection{Semantic SLAM}
SLAM algorithms allow robots to localize themselves while building a map of their surroundings. While SLAM primarily focuses on geometric reconstruction, recent variants provide impressive tracking and reconstruction abilities. Several prior systems give robots the ability to model the semantic meaning of the objects in their environment.  Many works used explicit scene representations, such as volumes, point clouds, scene graphs, and truncated signed distance functions (TSDF), for semantic scene understanding. \citet{mccormac2018fusion++} proposes an object-oriented online SLAM system to produce semantically labeled TSDF object instances reconstructions and 3D foreground masks. Object segmentation~\cite{dai2017scannet,engelmann20203d,hou20193d,lahoud20193d,qi2017pointnet++} and detection~\cite{qi2019deep} have been leveraged on volume and points clouds for semantic scene understanding.~\citet{rosinol2020kimera} provides a library for semantic SLAM by constructing 3D semantic meshes from semantic hand labeled images. 3D scene graph has also been used to describe a static scene either on a given mesh~\cite{armeni20193d} or on the incoming frames of observations~\cite{wu2021scenegraphfusion,hughes2022hydra}.~\citet{wu2021scenegraphfusion} incrementally fuses the semantic prediction of the current observation into a global semantic graph.~\citet{hughes2022hydra} constructs a 3D scene graph incrementally based on topological maps of locally built Euclidean Signed Distance Functions (ESDFs). 

Recent research in open vocabulary vision models~\cite{li2022languagedriven, ghiasi2022scaling, liang2023open, kirillov2023segany, xu2022odise} leads to increasing interest in creating 3D representations that incorporate semantic features. 
To reconstruct a static semantic SLAM, ConceptFusion~\cite{conceptfusion} combines SLAM and semantic features by projecting CLIP~\cite{radford2021learning} and features into 3D, where the labels are refined by 2D unsupervised segmentation. 

\subsection{Dynamic Scene Representation}
Environments often undergo changes over time. Extensive study has been done on detecting 2D scene changes~\cite{park2021changesim,ru2020multi,daudt2018fully}. For 3D environments,~\citet{rosinol20203d} represents dynamic scenes with moving agents with 3D dynamic scene graphs by integrating object and human detection and pose estimation model.~\citet{looper20233d} uses the Variable Scene Graph (VSG), an augmentation of 3D scene graph, to represent semantic scene changes. The variability of VSG is estimated in a supervised way. Prior work~\cite{schmid2022panoptic,finman2013toward,fehr2017tsdf,langer2020robust} has shown success using TSDFs in capturing long-term changes. Other works have used 2D point cloud data~\cite{lazaro2018efficient} or spatiotemporal grid~\cite{krajnik2016persistent} to learn the long-term environment changes but don't provide semantic understanding. These systems also require huge storage overhead.

Neural Radiance Fields (NeRF)~\cite{mildenhall2021nerf} are attractive alternative representation for high-quality scene reconstruction from pose RGB images, with an explosion of recent work on visual quality~\cite{adamkiewicz2022vision,barron2021mip,barron2022mip,ma2022deblur,huang2022hdr,sabour2023robustnerf,philip2023radiance}, large-scale scenes~\cite{tancik2023nerfstudio,wang2023f2,barron2023zip}, optimization speed~\cite{muller2022instant,chen2022tensorf,fridovich2023k,yu2021plenoxels}, dynamic scenes~\cite{park2021hypernerf,li2023dynibar,pumarola2020d}, combining SLAM and NeRF~\cite{sucar2021imap, zhu2022nice, lisus2023towards, rosinol2022nerf, chung2023orbeez} and more. 

Most similar to this work are methods which seek to adapt NeRF to a continual framework, such as Evo-NeRF~\cite{kerr2022evo}, ICNGP~\cite{Po2023InstantCL}, \citet{yan2021continual}, CLNeRF~\cite{cai2023clnerf}, and \citet{fu2022ndfchange}. However, prior works either don't consider changing scenes or naively discard stale images. Note that this goal is different from dynamic NeRF reconstruction~\cite{liu2023robust,karaoglu2023dynamon}, which must also recover the full scene motion, and 3D inpainting~\cite{yin2023or}, which must hallucinate the scene without an object. In~\algabbr{}, we leverage DROID-SLAM~\cite{teed2021droid} to obtain camera poses. We then actively detect changed regions and selectively update them to avoid unnecessarily discarding images. 

% In \algname{} we leverage the recent development of NeRF to construct implicit neural representations of scenes. A few recent studies have tackled the problem of cooperating implicit neural representations and visual SLAM, using an MLP~\cite{sucar2021imap} or a multi-level feature grid~\cite{ zhu2022nice, lisus2023towards} for scene representation. By leveraging the output from dense monocular SLAM, ~\citet{rosinol2022nerf, chung2023orbeez} show NeRF construction as the scene representation in real time from only monocular images.

% Similar to~\cite{rosinol2022nerf}, we obtain the camera poses from dense monocular SLAM (DROID-SLAM). Instead of constructing a NeRF directly from the estimated camera poses, we instead train a LERF, which encodes semantic information from CLIP along with the 3D geometry reconstruction. To capture scene changes, we use the difference in pretrained image features CLIP features, without the usage of object detection or segmentation models. This 

\subsection{Cloud and Fog Robotics}
Cloud Robotics~\cite{kehoe2015survey} is an emerging computational paradigm for robotics that enables robots with limited onboard computing capability to gain on-demand computing hardware resources. 
Exemplary cloud robotics applications include SLAM, motion planning~\cite{ichnowski2020fog} and grasp planning~\cite{tian2017cloud, li2018dex}.
% rapyuta, robocore, aws greengrass, google cloud robotics core
Major Cloud service providers, such as Amazon Web Services (AWS) and Google Cloud Platform (GCP), provide proprietary interfaces for robotics applications to interact with cloud interfaces, such as AWS Greengrass~\cite{greengrass} and Google Cloud Robotics Core.  
Rapyuta~\cite{mohanarajah2014rapyuta} is a platform for centralized management and deployment of a pipeline of robotics applications. 
% fogros 1 and 2, automatically launch the cloud instance 
FogROS~\cite{chen2021fogros} is the first open-source cloud robotics platform that offloads robotics applications to public cloud. FogROS enables robots to interact with the Cloud through the Robot Operating System (ROS) interfaces. 
\citet{ichnowski2022fogros} present FogROS2 that supports ROS2 and more major cloud service providers.
% fogros sgc, extend the connectivity, and remove the constraint of connecting only to the public cloud services 
FogROS2-SGC~\cite{chen2022fogros, chen2023fogros} secure and globally connects distributed robots with a peer-to-peer network. 
% In this work, we extend FogROS2 to a low-powered and inexpensive mobile robot.
In this work, we apply FogROS2 to a low-powered and inexpensive mobile robot.

% \subsection{NeRF in Robotics}
% {\color{blue}{is this redundant to the last para of section A?}}
% Neural Radiance Fields (NeRF)~\cite{mildenhall2021nerf} are an attractive representation for high-quality scene reconstruction from pose RGB images, with an explosion of recent work on visual quality~\cite{adamkiewicz2022vision,barron2021mip,barron2022mip,ma2022deblur,huang2022hdr,sabour2023robustnerf,philip2023radiance}, large-scale scenes~\cite{tancik2023nerfstudio,wang2023f2,barron2023zip}, optimization speed~\cite{muller2022instant,chen2022tensorf,fridovich2023k,yu2021plenoxels}, dynamic scenes~\cite{park2021hypernerf,li2023dynibar,pumarola2020d}, and more. Most similar to this work are methods which seek to adapt NeRF to the continual setting, such as Evo-NeRF~\cite{kerr2022evo}, ICNGP~\cite{Po2023InstantCL}, \citet{yan2021continual}, and CLNeRF~\cite{cai2023clnerf}. However, prior works either don't consider changing scenes or naively discard stale images, while in this work we actively detect changed regions and selectively update them to avoid unnecessarily discarding images.
 %P1-2
\section{Problem Statement}\label{sec:ps}

\begin{figure}[t]
\centering
\includegraphics[width=\linewidth]{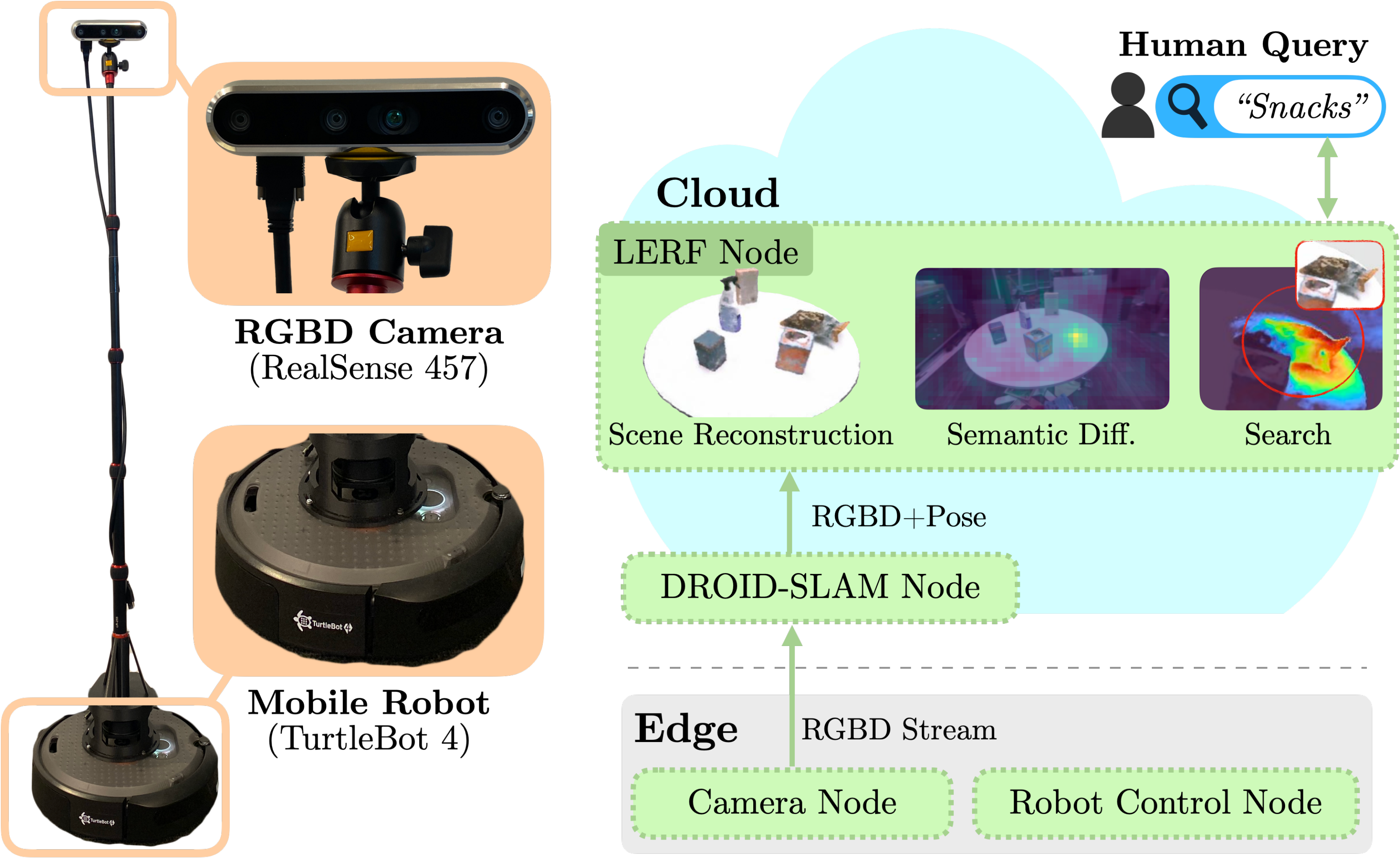}
\caption{\emph{Experiment setup and FOG-ROS2 Integration.} \textbf{Left}: We use a TurtleBot 4 as the mobile robot. We mount a RealSense D457 RGBD camera on top of the robot via a monopod. \textbf{Right}: We use FOG-ROS2 to execute both DROID-SLAM and LERF on a cloud machine. In particular, after the DROID-SLAM node obtains paired RGB and depth observations, it computes the camera pose. The LERF node reconstructs the LERF and calculates if the new observation is semantically inconsistent with the stored representation.
% We use the NeRF~\cite{mildenhall2021nerf, nerfstudio} coordinate convention for cameras: +X is right, +Y is up, and +Z is pointing back and away from the camera.
}
\label{fig:setup}
\end{figure}

For our algorithm, we assume that:
\begin{enumerate}[(i)]
    \item Objects are distributed on a local flat surface with sufficient visibility during a trajectory surrounding them.
    \item During any scan of the scene, the objects in the scene are static.
    % \item 
\end{enumerate}
% 1) Objects are distributed on a flat surface with sufficient visibility during a trajectory surrounding them. 2) During any scan of the scene, the objects in the scene are static.

We consider a typical SLAM scenario, with a wheeled, mobile robot equipped with an RGBD camera on a pole mounted on the robot. We consider several time periods: in the first time period, the robot constructs an initial model of the static environment. Between subsequent time periods, a set of objects may have been moved, added, or removed, and the goal of the robot is to detect these unknown changes and update the model accordingly. After each time period, the robot should be able to localize objects specified through natural language queries, even when objects are moved between time periods.

% \textbf{Problem Setup}
% A human teleoperator drives the mobile robot shown in Fig.~\ref{fig:setup} to generate a complete scan, from which a LeRF~\cite{lerf2023} is constructed and trained till convergence. The robot's start pose and trajectory are recorded and the trajectory is segmented by human-defined keyframes \todo{do we need this here?}. The entire trajectory or a partial trajectory starting from a keyframe can be replayed without running into obstacles. At any time step $t$, we have access to the RGB observation $o(t)$, and depth observation $d(t)$, from which we estimate the camera pose $p(t)$. 
 %P2-3
\section{Method}\label{sec:method}
\algabbr{} builds a LERF while the robot moves, updates it through environment changes, and localizes natural language queries. It has three main modules: 
\begin{enumerate}
    \item Scene reconstruction takes in a stream of RGBD images and executes DROID-SLAM, concurrently building a LERF with camera poses from SLAM and RGB image observations.
    \item The semantic differencing module compares incoming scene observations to the current reconstruction, producing 3D bounding boxes around changed regions.
    \item The LERF updating module takes in these bounding boxes and masks stale regions of images to ensure the reconstruction adapts to the new scene.
\end{enumerate}
% 1) Scene reconstruction takes in a stream of RGBD images and executes SLAM, concurrently building a LERF with these poses. 2) The semantic differencing module compares incoming scene observations to the current reconstruction, producing 3D bounding boxes around changed regions. 3) The LERF udpating module takes in these bounding boxes and masks stale regions of images to ensure the reconstruction adapts to the new scene.

\subsection{Constructing LERF}
\subsubsection{Camera pose estimation}
To construct the LERF representation, we need accurate camera poses from an RGBD camera stream. To provide these, we use DROID-SLAM~\cite{teed2021droid} to estimate camera poses. DROID-SLAM is a hybrid deep learning and optimization-based SLAM system that takes in monocular, stereo or RGB-D video and outputs per-keyframe disparity maps and keyframe poses. It iteratively updates predicted camera poses and pixel-wise depths through a Dense Bundle Adjustment layer, a differentiable module that computes a Gauss-Newton update to camera poses and pixel-wise depth. We choose to use DROID-SLAM for its strong pose tracking performance, and also for its ability to function without IMU and at lower camera fps (10hz), making it more robust to network latency. 
 
We directly feed the output keyframes and poses into the LERF to use as input views. We first initialize the pose prediction process with 4 RGBD images to ensure DROID-SLAM's scene scale matches physical scale, then switch to pure RGB estimation, which we find qualitatively drifts less. After adding initial poses, camera poses are additionally refined over the course of NeRF optimization by Nerfstudio's camera optimization~\cite{tancik2023nerfstudio}, a process which uses the differentiable rendering capabilities of NeRF to fine-tune poses, as proposed in BARF~\cite{lin2021barf}.

\subsubsection{Scene box determination}
Camera poses generated by DROID-SLAM are not necessarily axis-aligned with real-world coordinates, primarily because they are determined through an optimization process that aims for internal consistency rather than alignment with an external coordinate system. However, the hashgrid encoding~\cite{muller2022instant} assumes all cameras fit within a pre-defined scene box, an important step for ensuring parameters are sufficiently distributed throughout the scene. To account for this, we automatically set the scene scaling based on the first time period of mapping after the whole scene has been traversed.

\subsubsection{Concurrency in LERF Computation}
Training for LERF typically takes minutes to complete, which is not sufficient in this context as the robot must actively detect environmental changes and decide whether to update the underlying LERF model. In the standard LERF training process, the input images are first pre-processed by extracting CLIP and DINO features, a process that takes minutes. To enable continual LERF optimization, we introduce two modifications to the LERF training process. First, we continuously optimize the LERF model as the robot scans the scene, dynamically adding images during training as opposed to training on a fixed set. Second, we compute CLIP and DINO features in parallel with the ongoing LERF optimization, computing features in a callback and offloading completed images lazily within the train loop. This requires significant computation, about 500ms on a 4090 GPU per image, motivating the usage of cloud robotics to offload this process.

\subsection{Scene Change Detection}
\label{sec:method-diff}
\begin{figure}
    \centering
    \includegraphics[width=\linewidth]{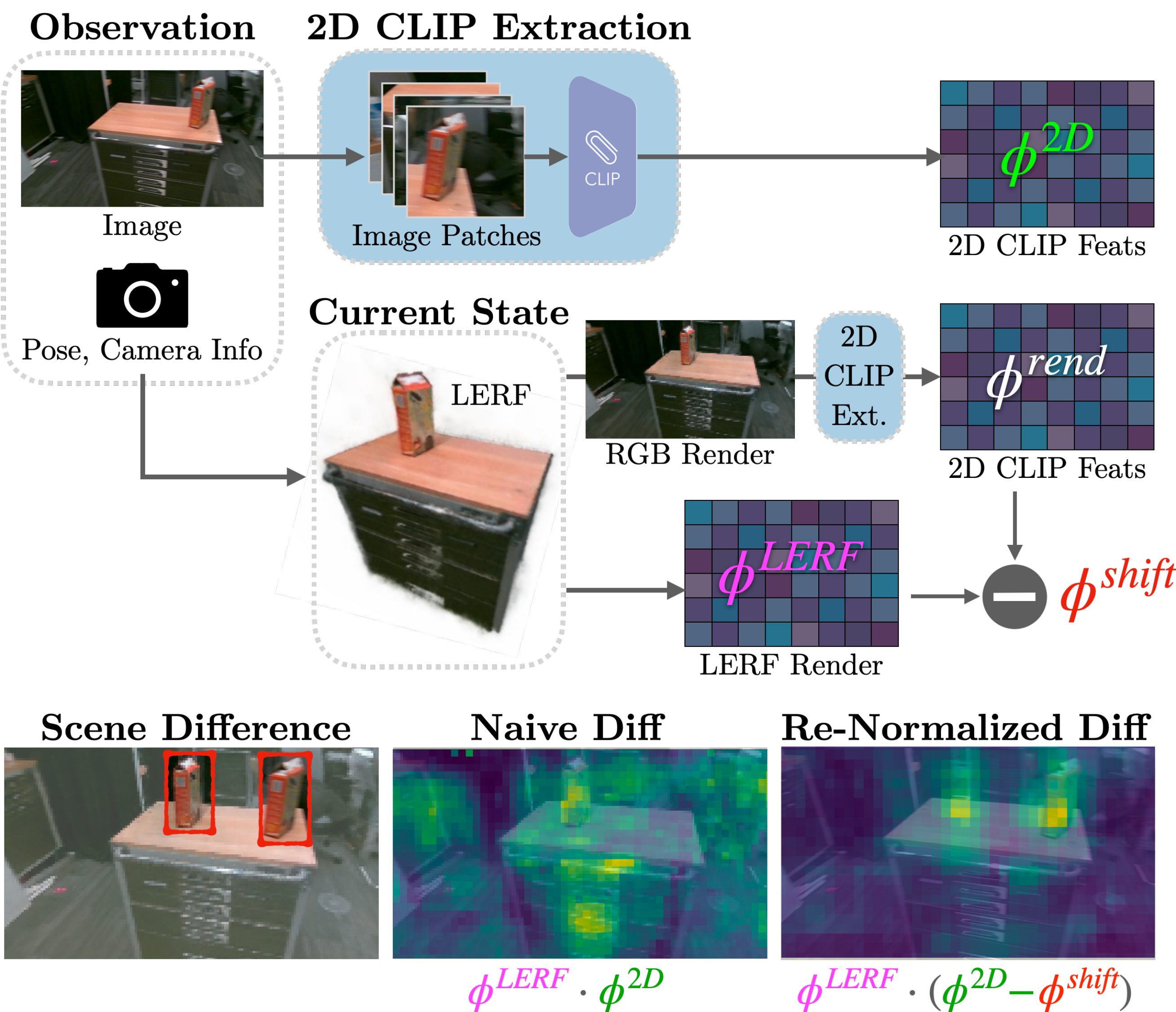}
    \caption{\textit{Semantic differencing}. The semantic differencing module calculates 2D feature maps from the fresh observation (top), the 3D LERF embeddings (middle), and the 2D CLIP embeddings of a NeRF-rendered image. $\phi^{rend}$ approximates the distribution shift from 2D to 3D, resulting in higher-quality semantic difference heatmaps (bottom). See Sec~\ref{sec:method-diff} for details.}
    \label{fig:sem-diff}
\end{figure}
Motivated by the constantly changing environments in homes and warehouses, \algabbr{} detects changes in the environment over time so that it can selectively update these regions. Given an image and approximate camera pose as input, the output of this module is a set of 3D bounding boxes of changed regions. We explore a semantic differencing method based on language embeddings rendered from CLIP for computing 3D changed regions. This method leverages the novel view synthesis ability of implicit neural fields for comparing the current reconstruction to new input views. 

% \subsubsection{DINO Attention Maps}
% DINO~\cite{dino} is a self-supervised pretraining framework that offers Visual Transformer (ViT) features. These features encapsulate explicit information related to the semantic segmentation of images. Similar to the unsupervised object segmentation approach used in DINO~\cite{dino}, to identify changes in a scene, a forward pass is executed using the DINO ViT-S model on both the newly acquired image from the robot and a corresponding image rendered by LERF, based on the camera pose of the new image. We calculate the pixel-wise maximum across all eight attention heads and normalize the resultant feature map to a range between 0 and 1. A mask is then obtained for each model by applying a threshold of 0.4. Changes in the scene are ascertained by taking the union of the differences between these segmentation masks.
\subsubsection{3D LERF Feature Difference}
Since the primary use case of a semantic map is for semantic queries, this approach uses the underlying language embeddings to identify scene changes. 
% This way, updates are only carried out when they have a functional effect on the downstream task at hand. 
To detect semantic changes, we first render the 3D field from the same camera pose as the new image to obtain a 2D feature map $\phi^{LERF}$. Since LERF is a multi-scale representation, we pick an image scale of 0.25 for querying the underlying field with and use the same scale to generate 2D embeddings by sliding a crop over the input image and passing the crops through the CLIP image encoder to obtain $\phi^{2D}$. 

Naively computing the difference $\phi^{LERF}-\phi^{2D}$ falsely activates on edges of the scene. This is because the CLIP embeddings in 3D experience a distribution shift from their respective input views. To understand this, consider a scene with a tall narrow object: input views of this object contain primarily the background so their CLIP embeddings are dominated by background features. However, the background is inconsistent between different viewing angles, meaning the average embedding in 3D will be different than each 2D view. To overcome this, we propose a method for estimating this 2D-to-3D distribution shift and apply it to obtain a \textit{semantic re-normalization} of differences obtained. Let $\phi^{rend}$ be the 2D grid of embeddings obtained by calculating CLIP embeddings of sliding a crop window over a \textit{rendered} RGB image from NeRF. Then, the distribution shift can be quantified as $\phi^{LERF}-\phi^{rend}$. We subtract off this shift from the 2D image embeddings, re-normalize, then compute the dot product with 3D embeddings to find the difference. Mathematically, this is $$\texttt{renorm}(\phi^{2D}-(\phi^{LERF}-\phi^{rend})) \cdot \phi^{LERF}$$
This re-normalization is critical; visualizations of an ablation are provided in Fig.\ref{fig:sem-diff}.

\subsubsection{3D Box Detection}
We utilize the semantic differencing module while the robot drives to estimate 3D changed regions. To do this, we binarize heatmaps with a threshold of 0.9, with values below 0.9 cosine similarity registering as different. We then run Tarjan's connected component's algorithm~\cite{tarjan1972depth} to parse the heatmap to discrete heatmaps that do not intersect. We then filter out heatmaps with less than 30 pixels and deproject each heatmap into 3D by using the minimum of NeRF depth and RealSense depth, to account for the potential of adding or removing objects. We concatenate point clouds from 15 images at a time, then compute clusters using DBSCAN and fit oriented bounding boxes to each cluster. We constrain boxes' Z axes to align with the world axis.
\subsection{Updating LERF}
Once changes have been identified in 3D, we must selectively update both the language and visual properties of the LERF to match the scene state. To do this, we mask input NeRF views from the previous stage by projecting the 3D bounding boxes into each training view. We then prevent rays from being sampled inside image masks, which effectively prevents the LERF from incorporating information from stale regions of images. During each time period of exploration, we only mask images from prior time periods to avoid over-masking images. NeRF leverages \textit{proposal networks}~\cite{barron2022mip} to guide the volumentric sampling process, which represents a low-frequency 3D distribution of where geometry lies in the scene. We modify the proposal sampling procedure by adding a constant probability mass of 0.02 to each ray sample, determined empirically, ensuring that samples are always taken in free space. This is important for convergence in locations where objects are added as otherwise the field would never be sampled where objects are added.

\subsection{Fog-ROS2 Integration}
% as a result, we integrate \algname with FogROS2. we package LERF is packaged into ROS nodes and pass images through pub/sub. We connect the robot and cloud with the proxy of FogROS2 that can conenct with secure and global connectivity 
Wheeled mobile robots possess limited computation power designed for basic controllers and sensors. Consequently, they cannot execute \algabbr{} on their single-board computer, while both DROID-SLAM and LERF necessitate distinct GPUs. 
To overcome this limitation, we incorporate Fog-ROS2 to facilitate access to on-demand cloud computation via a separate workstation. 
We encapsulate DROID-SLAM and LERF in ROS2 nodes and
implement the publish/subscribe ROS2 interfaces; consequently, the robot can publish images to DROID-SLAM in the cloud, which consecutively publishes the maps and keyframe poses with LERF. 

%   . In this work, we implement an adaptive algorithm that controls the speed of how the proxy spins with how fast the message comes in, to enable the availability with high QoS 
% We observe that the network routing framework of FogROS2 requires substantial CPU resources, continually operating for optimized message responsiveness, while the CPU resource on the battery-powered mobile robots is already dedicated to sensors and controllers. Straightforward limiting the CPU usage of FogROS2 results to packet drops and misaligned camera frames and pose. We address this issue by designing an adaptive algorithm that reduces message processing frequency while maintaining the quality of service. We design a multiplicative increase additive decrease rate controller that dynamically adjusts the transmission rate, incrementing it multiplicatively during favorable conditions and decreasing it additively in response to higher CPU usage. The system stabilizes quickly, because \algabbr{} runs with fixed number of frames per second.  

% \todo{white space reserved for todos in method}
% \newpage  %P3-4
\section{Experiments}\label{sec:experiments}
To evaluate \algname{} we consider a local table with up to 5 objects lying on it (see Fig.~\ref{fig:setup}). We use a Turtlebot 4 equipped with a RealSense D457 camera which faces toward the starboard side. SLAM and LERF optimization executes on a workstation with 2 RTX 4090s (Fig.~\ref{fig:setup}). This direction is chosen to maximize camera parallax to enhance SLAM pose estimation and NeRF quality. In addition, it makes mapping inward-facing scenes easier with a differential drive base. Experiments progress in \textit{time periods}: we first capture the entire scene by driving the robot in a fixed trajectory around the table and train the LERF for 2000 steps (2 minutes) before initiating the next time period. All consecutive queries happen without additional idle training time, only updating the LERF during robot motion.
\begin{figure}[t]
\centering
\includegraphics[width=\linewidth]{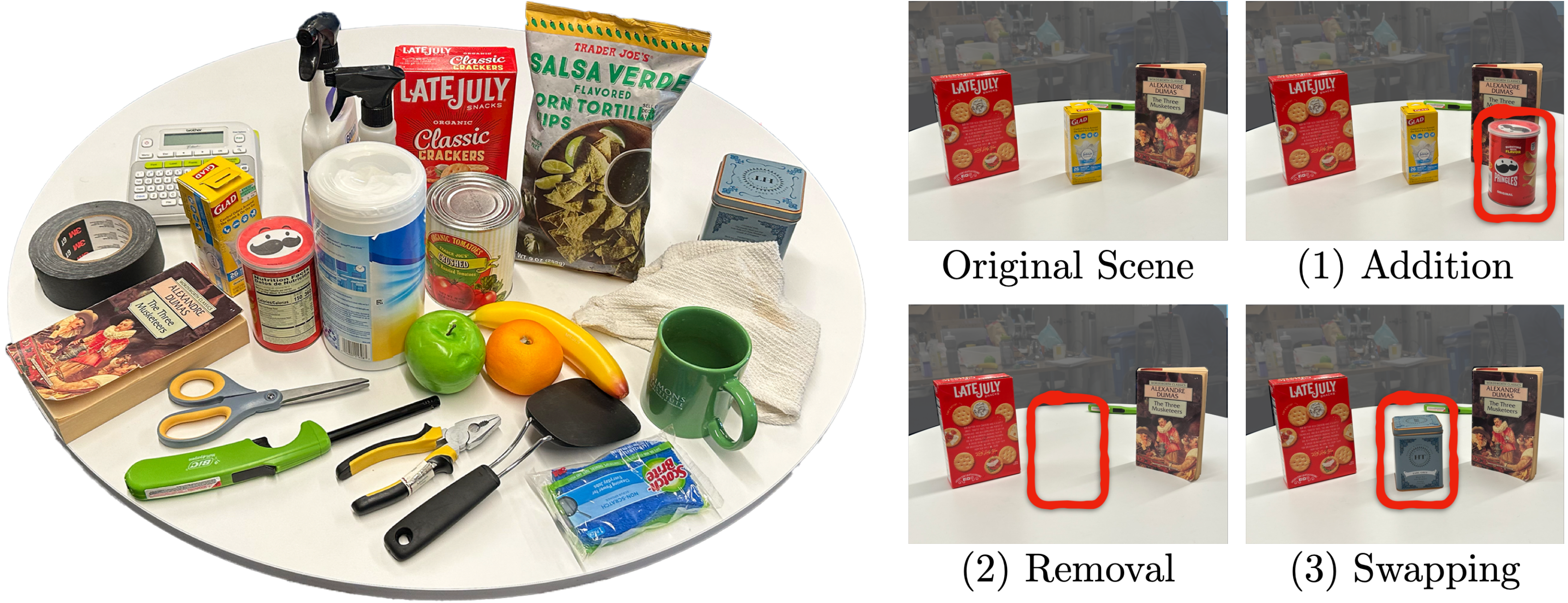}
\caption{\textbf{Experiment setup} \textit{(Left)}: Test objects; (\textit{Right}): Three types of scene changes included in evaluation. Red box denotes the changed scene region.}
\end{figure}

\begin{figure}
\centering
\includegraphics[width=\linewidth]{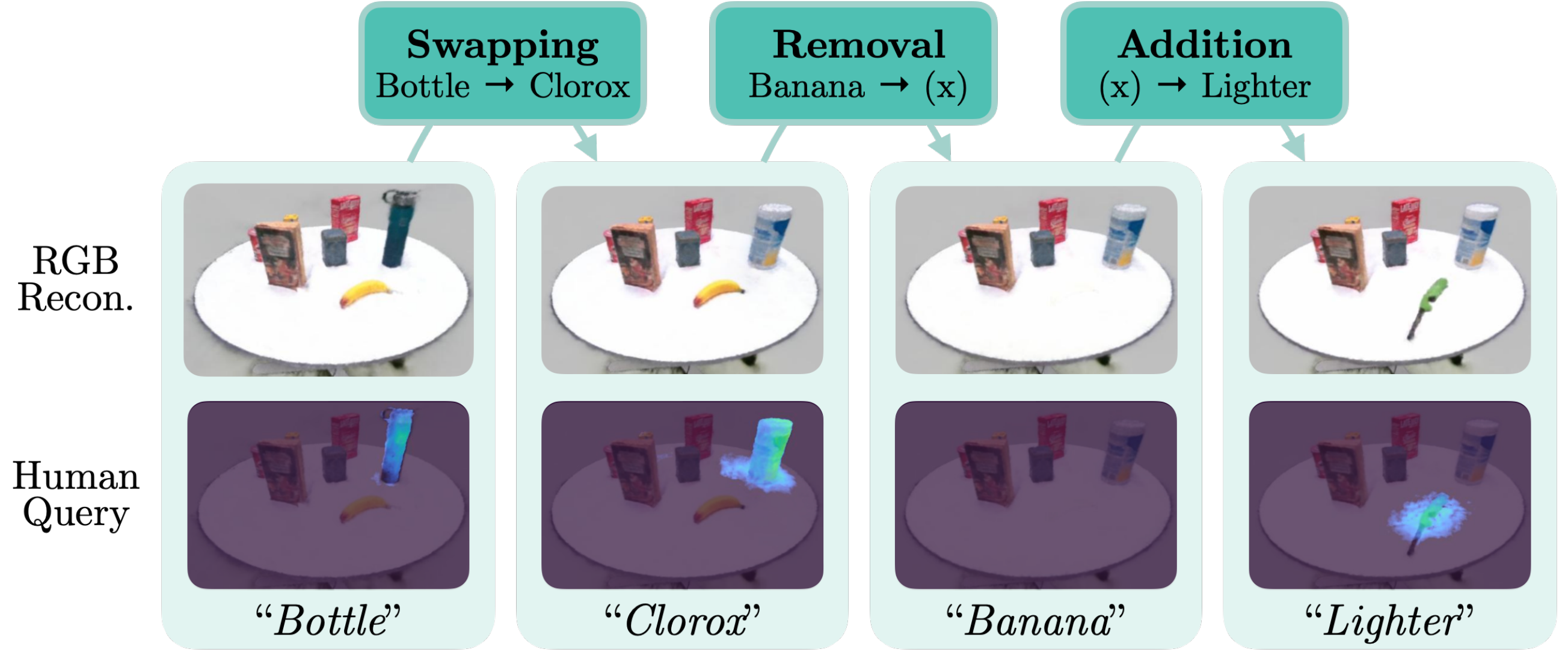}
\caption{\textit{Sequential scene update.} Scene reconstructions are shown in the middle, and human queries are shown at the bottom. As the scene updates, the heatmap for human queries updates accordingly.}
\label{fig:seq_update}
\end{figure}

In subsequent time periods, we randomly select one or two objects to add, remove, or replace within the scene (Fig.~\ref{fig:seq_update}), and the robot executes a partial trajectory to detect changes and decide whether to remap or not. If it detects changes, it completes its trajectory around the table and adds new images to the LERF, otherwise it early-terminates. Between each time period, a language query for each object is sent to the LERF; both updated and static objects. A language query is considered a success if the argmax language activation in the 3D scene lies on the correct object. For objects that were removed, we consider the localization a success if the argmax language activation no longer lies on its previous location,
% no objects have positive language activation, 
and for swapped objects if a query for the removed object no longer activates on swapped in object. LERF is well known to struggle with answering ``existence" questions~\cite{lerf2023}, so to evaluate how well the language embeddings are replaced we ensure removed objects' language queries no longer remain in the same location, and maintain the same activation as the background.

The experiments divide possible changes in a scene into 4 categories:
\begin{enumerate}
    \item \textit{No Change}: 
    after the initial scan, no objects are removed from the scene. 
    % the initial scan is the same as the second scan of the scene, where no object is moved or changed. 
    This measures a scene difference detector's bias towards false positive prediction (i.e. model predicting that the scene is changed and needs to be remapped but instead is unchanged).
    \item \textit{Removal}: after the initial scan, one or two objects are removed from the scene. 
    \item \textit{Addition}: after the initial scan, one or two objects are added to the scene. 
    \item \textit{Swap}: one or two objects are placed in the scene after the initial scan.
    % \item \textit{Moving}: Existing objects' positions can be moved 
    % \item \textit{Full rearrangement}: A combination of the above
\end{enumerate}

For every experiment with scene changes, we execute 2 difficulty tiers per category and 2 trials per tier. Tier 1 moves 1 object between each time period, and Tier 2 moves 2 objects between each time period. All experiments consist of 3 time periods, except the no change experiment which has 2.

During experiments for add, remove, and swap we measure
% 1) Peak Signal-to-Noise Ratio (PSNR) between new input views and the prior input views to evaluate how well the visual reconstruction adapts 
1) the percent of changed objects detected by \algabbr{}, defined by outputting bounding boxes fully containing them, 
% 2) the amount of false positives the system outputs, tested by mapping the same scene twice 
2) the number of pixels masked in the original image, and
3) language query accuracy for each object on the table after each time period. We aim for recall to be as close to 100\% while minimizing the number of masked pixels, preserving the original input images wherever possible. We separately calculate language query accuracy for moved and static objects, to separately evaluate the robustness of the LERF updating method for adapting language embeddings without affecting existing regions of the scene. Finally, to test the false positive rate of the system, we execute the same trajectory twice and measure whether the system falsely detects a change in the scene. 

%To maintain fair comparison between differencing methods and facilitate testing on different clouds, we use ROS to bag the raw sensor data and replay it in real time during experiments.

\begin{table*}[]
\centering 
\resizebox{0.9\textwidth}{!}{%
\begin{tabular}{rllllll}
\hline
\multicolumn{1}{l|}{}                  & \multicolumn{2}{c|}{Add}        & \multicolumn{2}{c|}{Remove}     & \multicolumn{2}{c}{Swap} \\
\multicolumn{1}{r|}{Change Detection Method} & Baseline & \multicolumn{1}{l|}{\algname{}} & Baseline & \multicolumn{1}{l|}{\algname{}} & Baseline & \algname{} \\ \hline
Language Query Accuracy (Moved Objects) (\%)                  & 75    & \textbf{92}                        & 89    & \textbf{90}                        & 67    & \textbf{92}       \\
Language Query Accuracy (Static Objects)(\%)                  & 83    & \textbf{90}                        & 97    & 97                        & 63    & \textbf{75}       \\\hline
Pixel Mask Ratio (\%)                  & \textbf{13}    & 21                        & 20    & \textbf{17}                        & \textbf{17}    & 24       \\\hline
Change Detection Recall (\%)                    & 92    & 92                        & 83    & \textbf{92}                       & 67    & \textbf{92}       \\ 

\end{tabular}
}
\caption{\textbf{Results}: We evaluate the two scene change detection methods across 8 scenes. \algabbr{} has on average an 91\% language query on accuracy objects moved throughout the trial, highlighting its ability to adapt both geometry and language as the scene changes. Both depth and semantic differencing can detect scene changes with high recall with similar pixel mask ratios, but semantic differencing is more robust against false positives (see Tab~\ref{tbl:false_pos}).}
\label{tbl:main_result}
\end{table*}

\subsection{Baseline: Depth-Camera Image Differencing}
We compare against depth-based image differencing where the robot renders depth from the stored LERF at the same pose as the incoming image and calculates the pixel-wise absolute difference between the two depth images, then filter differences which are larger than 30cm or smaller than 10cm, and rejects depth values outside the range 10cm to 150cm. The resulting binary heatmap is passed through the same pipeline described in Sec~\ref{sec:method-diff}. Note that both the baseline and lifelong LERF models have the same language query time and change detection time, since they share the same LERF architecture and make the same number of 2D renders.

% To account for sensing errors (i.e., occlusion, where depth will be zero in objects' shadows, and larger errors for background objects), we only consider pixels that are between 0.1 and 1.5m. To account for small sensing errors due to stereopsis and sensor noise, we only consider absolute pixel depth differences that range from 0.1 to 0.3m.Sec.

% 1. Comparing success in detection of changes in a scene (focus on false positive), with varying level of difficulties \\
% Scene: Cluttered scene of 15-20 objects on a table / shelf. 
% Stages:
% 1. Object Addition: Add objects to the scene, without perturbing the layout of the items within the original scene
% 2. Object Removal: Remove objects from the scene, without perturbing other objects not being removed from the scene
% 3. Object Addition \& Removal: Do a combination of object additions and removals. Potentially swap two objects (removal \& addition in the same location)
% 2. Success rate in creating LeRF from droidslam pose estimates (on static scenes) + computation time \\
% 3. Success rate in updating LeRF based on droidslam pose estimates + computation time (vs training from scratch) [maybe should merge with 1?]
\section{Results}\label{sec:results}
% \begin{table*}[]
% \centering
% \resizebox{\textwidth}{!}{%
% \begin{tabular}{rllllllllllll}
% \hline
% \multicolumn{1}{l}{}    & \multicolumn{3}{c}{No Change} & \multicolumn{3}{c}{Add} & \multicolumn{3}{c}{Remove} & \multicolumn{3}{c}{Replace} \\
% Algorithms              & RGB     & Depth     & CLIP    & RGB   & Depth   & CLIP  & RGB    & Depth    & CLIP   & RGB    & Depth    & CLIP    \\ \hline
% Detection Accuracy (\%) & 50      & 60        & 70      & 50    & 60      & 70    & 50     & 60       & 70     & 50     & 60       & 70      \\
% Detection Recall (Move)(\%)   & 50      & 60        & 70      & 50    & 60      & 70    & 50     & 60       & 70     & 50     & 60       & 70      \\
% Time (s)                & 10      & 20        & 30      & 10    & 20      & 30    & 10     & 20       & 30     & 10     & 20       & 30      \\
% Language Query (\%)     & 40      & 50        & 70      & 40    & 50      & 70    & 40     & 50       & 70     & 40     & 50       & 70      \\ \hline
% \multicolumn{1}{l}{}    &         &           &         &       &         &       &        &          &        &        &          &         \\
% \multicolumn{1}{l}{}    &         &           &         &       &         &       &        &          &        &        &          &         \\
% \multicolumn{1}{l}{}    &         &           &         &       &         &       &        &          &        &        &          &        
% \end{tabular}
% }
% \caption{Table to test captions and labels.}
% \label{tbl:main_result}
% \end{table*}

\begin{table}[]
\centering 
\resizebox{.8\linewidth}{!}{%
\begin{tabular}{rll}
\hline
\multicolumn{1}{l|}{}           & \multicolumn{2}{c}{No Change}          \\
\multicolumn{1}{r|}{Change Detection Method} & Baseline & \algname{} \\ \hline
Decision Accuracy             & 0\%    & \textbf{80\%}   \\ 
Pixel Mask Ratio              & 8\%    & \textbf{1.2\%}   \\ \hline
\end{tabular}
}
\caption{\textbf{False Positive Evaluation}: Depth over-predicts due to SLAM pose drift and NeRF frame misalignment, leading to high false positives. Lifelong LERF is more pose-error tolerant.}
\label{tbl:false_pos}
\end{table}
We average the detection percent and the localization accuracy across 4 different scenes for each category of scene change and summarize the results in Tables~\ref{tbl:main_result},~\ref{tbl:false_pos}. Empirically, the depth-differencing baseline tends to mistakenly identify most objects in the scene as changed because of subtle camera pose drift over time. Any pose misalignment produces false positives in the depth difference and results in a heatmap that over-segments objects. 

\subsection{No Changes}
Because depth is sensitive to pose misalignment, the depth difference method falsely detects changes every time. \algabbr{} successfully predicts that there are no changes within the first 15 images 80\% of the time. This saves substantial time since the robot does not need to fully remap the environment. \algname{} is more robust to pose misalignment because this language difference is computed at a lower spatial frequency 
due to CLIP's~\cite{radford2021learning} receptive field.
% due to the receptive field of CLIP~\cite{radford2021learning}. 

% For scenes that are not changed, we found that RGB has a lot of false positives. Depth has less and is comparable with CLIP. Intuitively, this is because LERF rendering has large RGB differences before the model converges. 2D differencing is more susceptible to such changes in RGB; depth changes are less high frequency than changes in RGB so it is easier for LERF / NERF to learn, therefore there is less false positive. Semantic changes similar are regularized by geometry consistency, therefore are less susceptible to false positives. The number of images required for depth and RGB to find that the scene is unchanged and success rate for language queries for the two methods are similar due to the aforementioned reasons. 

\subsection{Adding and Removing}
When adding objects, the depth-differencing baseline averages 79\% language query accuracy and a 92\% change detection recall rate. \algabbr{} averages 91\% language query accuracy and detects changes in the environment correctly 92\% of the time. The depth-based baseline tends to under-select added objects, leading to insufficient updates to the LERF to update the 3D semantic features, resulting in slightly worse language performance. In the object removal setting, the depth-based baseline achieves comparable language query performance with~\algabbr{}. Prior work~\cite{kerr2022evo} has also demonstrated that it is comparatively easier to remove information from NeRF than to add new information. 
% For scenes where objects are added and removed, we found that RGB, depth and CLIP are comparable. Intuitively, 2D images are visually different, and depth and CLIP are also very different. Therefore, all methods gives high accuracy in determine the difference. However, RGB method is still more susceptable to changes in the scene, therefore is generating more bounding boxes than what there is in the scene. This results in a low recall of the method. (write something for depth and CLIP)

\subsection{Swapping Objects}
The depth baseline suffers significantly because swapping an object largely preserves scene geometry, while the semantic differencing method is able to capture the differences in the scene since the language embeddings between new and old objects differ significantly. This is reflected by the fact that the baseline localizes moved objects with 67\% accuracy, while semantic differencing localizes a 92\%.
% For scenes where objects are swapped, CLIP shows significantly better detection accuracy than depth, especially when swapped objects are geometrically (size and shape) similar to each other. For RGB baselines, as the heatmaps have high overlaps, and vanilla RGB difference gives activation everywhere due to camera pose error, it performs worse than the CLIP based method. Since the objects might be geometrically similar (reference figure 1 example of Clorox box vs cookie can), the resulting difference in depth is small, which lead to low detection accuracy. Since both RGB baseline and depth baseline are unable to detect objects, the success rate for language query is low because the LERF is not updated. 

 %P5-6
\begin{figure}
    \centering
    \includegraphics[width=\linewidth]{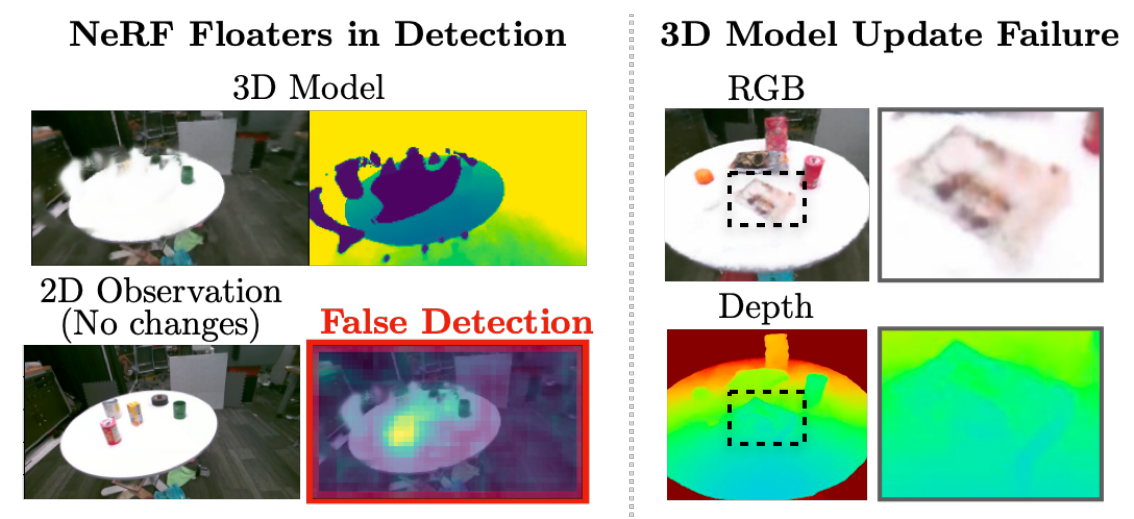}
    \caption{\textbf{\algname{} failure modes: } \textit{(Left): }NeRF's spurious density triggers false difference detection and alters CLIP feature differences through incorrect activations. \textit{(Right): }If semantic changes are not correctly detected, dataset inconsistency will form fuzzy density filled with holes and fail to fully add or remove the object.
    % \textbf{Failure due to floaters}: NeRF's spurious density triggers false difference detection and alters CLIP feature differences through incorrect activations.
    }
    \label{fig:limitations}
\end{figure}
\section{Limitations}
% \algabbr{} assumes that the scene is static while the robot is mapping, an assumption that would need to be relaxed for practical usage in dynamic scenes. In addition, this work uses pre-defined capture trajectories to map scenes; for large-scale deployment, autonomous exploration would be critical in addition to a more sophisticated camera setup with multiple outward-facing views. Semantic change detection can be sensitive to the size of objects, with smaller objects more difficult than large ones. Future work might study how to use multi-scale difference detection by aggregating across multiple levels. \algabbr{} has a tendency to over-segment changes in the scene, meaning that often too much of the scene is masked out. All of our experiments were run with a private cloud workstation. For practical deployment it is not feasible to own a powerful machine, so as a proof of concept, we integrated the system on an AWS cloud server with 3 A10G GPUs, at a cost of \$8/hr. Future work will study the real-time system performance associated with public cloud usage.

Some of the failure modes are presented in Fig.~\ref{fig:limitations}. \algabbr{} operates under the assumption of a known scene and uses pre-defined trajectories for mapping. It is not practical for highly dynamic environments and large-scale scenes, which may benefit from autonomous exploration and multi-camera setups. The algorithm also shows sensitivity to object size in semantic change detection and has a tendency to over-segment scenes. More sophisticated approach to difference detection which combines segmentation, semantics, geometry, and perhaps a learned difference detection module may be interesting future work.

\section{Conclusion}
In this study, we introduce \algname{}, a system that builds and updates a LERF to dynamically adapt to semantic changes in a scene. Using the dense semantics of LERF, we propose a method for detecting scene changes that is more robust to pose errors or reconstruction inaccuracy compared to a depth camera differencing baseline. Results suggest high accuracy in change detection with a minimal rate of false positives in local tabletop settings, and the ability to rapidly query the updated LERF with natural language to inventory objects. We offload computation to a local server using Fog-ROS2 to enable deployment on an inexpensive Turtlebot robot. Future work will study scaling this method to larger scale room environments, where more complex mapping is necessary.  %P6

% \section{Conclusions}

% , we plan to refine the CLIP feature extraction and comparison process for more nuanced change detection. We are also interested in evaluating \algname{} on diverse robot platforms to understand how computational offloading affects overall system latency and robustness. Finally, we aim to adapt our algorithm for use in dynamic environments where scene elements are subject to frequent changes.
% \section*{Acknowledgments}
% This research was performed at the AUTOLAB at UC Berkeley in affiliation with the Berkeley AI Research (BAIR) Lab. The authors were supported in part by donations from Toyota Research Institute, Bosch, Google, Siemens, and Autodesk and by equipment grants from PhotoNeo, Nvidia, and Intuitive Surgical.

\renewcommand*{\bibfont}{\footnotesize}
\printbibliography
\clearpage

\end{document}